\documentclass[letterpaper, 10 pt, conference]{ieeeconf}  
\let\labelindent\relax

\usepackage{amsmath}  
\usepackage{graphicx}
\usepackage{xcolor} 
\usepackage{enumitem} 

\usepackage{multirow}  
\usepackage{booktabs}  
\usepackage{float}     
\usepackage{arydshln}  

\IEEEoverridecommandlockouts                              
\title{\LARGE \bf
\textbf{MVS-GS: High-Quality 3D Gaussian Splatting Mapping via \\ Online Multi-View Stereo}
}

\author{Byeonggwon Lee$^{1}$, Junkyu Park$^{1}$, Khang Truong Giang$^{2}$, Sungho Jo$^{3}$*, Soohwan Song$^{1}$*%
\thanks{*Corresponding authors: {\tt\scriptsize songsh@dongguk.edu}, {\tt\scriptsize shjo@kaist.ac.kr}}
\thanks{$^{1}$Byeonggwon Lee, Junkyu Park, and Soohwan Song are with College of AI Convergence, Dongguk University, Seoul, Korea}%
\thanks{$^{2}$Khang Truong Giang is with 42dot, Seoul, Korea}%
\thanks{$^{3}$Sungho Jo is with School of Computing, KAIST, Daejeon, Korea}%
}

\begin{document}
\maketitle
\thispagestyle{empty}
\pagestyle{empty}

\begin{abstract}
This study addresses the challenge of online 3D model generation for neural rendering using an RGB image stream. Previous research has tackled this issue by incorporating Neural Radiance Fields (NeRF) or 3D Gaussian Splatting (3DGS) as scene representations within dense SLAM methods. However, most studies focus primarily on estimating coarse 3D scenes rather than achieving detailed reconstructions. Moreover, depth estimation based solely on images is often ambiguous, resulting in low-quality 3D models that lead to inaccurate renderings. To overcome these limitations, we propose a novel framework for high-quality 3DGS modeling that leverages an online multi-view stereo (MVS) approach. Our method estimates MVS depth using sequential frames from a local time window and applies comprehensive depth refinement techniques to filter out outliers, enabling accurate initialization of Gaussians in 3DGS. Furthermore, we introduce a parallelized backend module that optimizes the 3DGS model efficiently, ensuring timely updates with each new keyframe. Experimental results demonstrate that our method outperforms state-of-the-art dense SLAM methods, particularly excelling in challenging outdoor environments.
\end{abstract}


\section{INTRODUCTION}

Precise 3D models are in high demand for various industrial applications, including digital twins and virtual or augmented reality. One of the most commonly used methods for 3D modeling is \textit{multi-view stereo} (MVS) \cite{schonberger2016pixelwise} \cite{gu2020cascade} \cite{cao2022mvsformer}. MVS generates high-quality 3D models by identifying dense correspondences among multiple images taken from different viewpoints. Recently, MVS has been integrated with neural rendering techniques like \textit{neural radiance fields} (NeRF) \cite{mildenhall2021nerf} and has been expanded for use in novel view synthesis \cite{chen2021mvsnerf} \cite{liu2024geometry}. Specifically, 3D modeling using \textit{3D Gaussian splatting} (3DGS) \cite{kerbl20233d} \cite{zheng2024gps} \cite{liu2024fast} has made real-time, high-quality rendering possible. However, traditional MVS is designed as an offline algorithm, which requires a significant amount of time to process 3D modeling in batches. As a result, MVS has not been widely adopted in robotics or graphical applications that require real-time processing.

Dense \textit{simultaneous localization and mapping} (SLAM) \cite{newcombe2011dtam} \cite{pizzoli2014remode} \cite{koestler2022tandem} tackles the challenge of online 3D modeling. Typically, dense SLAM methods estimate depth maps using a motion stereo technique, performing MVS on sequential images within a local time window.
Recently, many studies have adopted NeRF \cite{chung2023orbeez} \cite{matsuki2023imode} \cite{rosinol2023nerf} \cite{zhang2023go} \cite{zhu2024nicer} \cite{zhang2024glorie} \cite{peng2024q} or 3DGS \cite{huang2024photo} \cite{matsuki2024gaussian} \cite{lan2024monocular} as map representations in dense SLAM, enabling real-time 3D modeling for rendering and view synthesis. However, existing methods primarily focus on estimating coarse 3D scenes rather than achieving detailed reconstruction. Most approaches rely on down-sampled images or lightweight networks for real-time computation, which significantly reduces the quality of the generated 3D models. Moreover, depth estimated solely from images is highly ambiguous due to factors such as motion blur, occlusion, or textureless regions. Despite this, current methods simply apply existing RGBD-based mapping techniques \cite{zhu2022nice} \cite{keetha2024splatam} that rely on reliable depth, often resulting in noisy reconstructions. Therefore, a new approach is needed to estimate high-resolution, accurate depth maps and incorporate stronger geometric priors for high-quality mapping.

\begin{figure}
    \centering
    \includegraphics[width=0.93\linewidth]{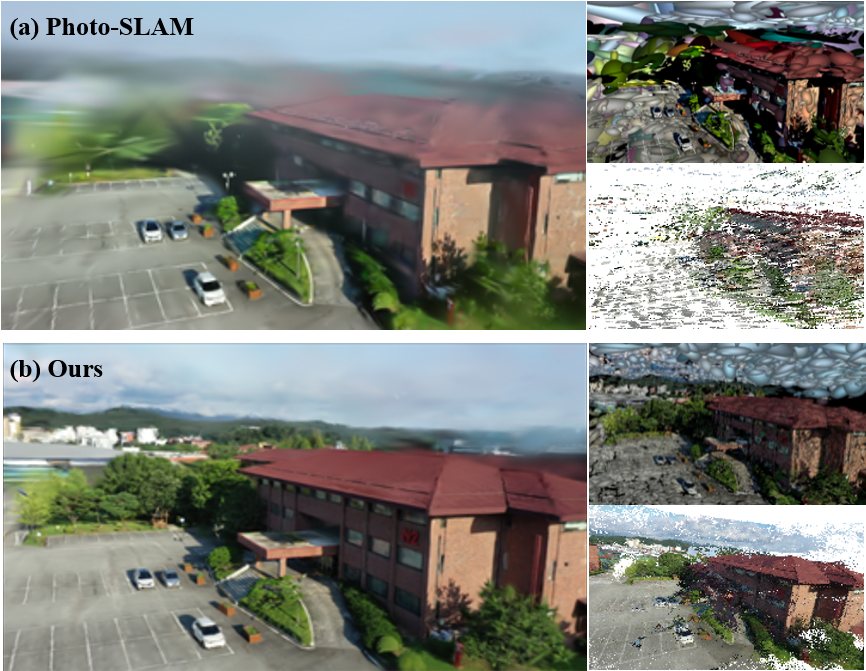}
    \caption{Rendering results of (a) Photo-SLAM \cite{huang2024photo} and (b) our method. The optimized Gaussian points and the estimated point cloud for each method are shown in the upper right and lower right, respectively. Our method utilizes an online MVS approach to generate a dense point cloud, which is then used to initialize Gaussian points. As a result, our method produces a 3DGS model with much higher rendering quality compared to Photo-SLAM, which relies on a sparse point cloud. }
    \label{fig1: rendering results}
    \vspace{-15pt}
\end{figure}

To address these challenges, we propose a novel framework that accurately estimates 3D information using an online MVS approach, which is then used to generate high-quality 3DGS models. The online MVS utilizes a sequence of frames within a local time window as source images and employs MVS networks \cite{cao2022mvsformer} to estimate the depth map of the current frame. This method enables the real-time generation of precise, high-resolution depth maps. Additionally, since depth estimates derived solely from images can contain numerous outliers and inaccuracies, we implement a comprehensive depth refinement and filtering process. This step improves the depth map by integrating sequentially estimated depth information, effectively removing outliers. Unlike existing methods \cite{huang2024photo} \cite{matsuki2024gaussian} \cite{lan2024monocular} that rely on sparse or noisy depths to generate 3DGS models, our approach produces much denser and more accurate depth maps, facilitating the initialization of detailed Gaussian points (see Fig. 1).

Additionally, we developed an efficient mapping framework to enhance the learning efficiency of high-quality 3DGS within a limited timeframe. This framework includes an independent backend module focused on optimizing 3DGS models, which runs in parallel with the frontend module responsible for camera tracking and depth estimation. This parallel setup ensures that the 3DGS has sufficient time to refine its parameters before the next keyframe is processed.
Finally, the proposed framework demonstrates superior rendering accuracy compared to existing dense SLAM methods and is particularly effective in complex outdoor scenes.

The key contributions of this work are as follows:

\begin{itemize}[leftmargin=0.5cm]
    \item We propose a novel framework for high-quality 3DGS mapping, leveraging an online MVS method. The precise depth maps produced by MVS facilitate the initialization of accurate and dense Gaussian points.
    \item We introduce an online MVS method that incorporates a comprehensive depth filtering process. This approach sequentially estimates depths from incoming frames, integrates them to produce temporally consistent depths, and effectively filters out outliers.
    \item An efficient backend for 3DGS mapping is designed to operate in parallel with the frontend, ensuring sufficient time for model optimization. This backend also reduces the number of particles by initializing Gaussian points only for unexplored regions.
    \item 
    The proposed framework has been evaluated on two benchmarks for indoor scenes \cite{straub2019replica} \cite{sturm2012benchmark}. Unlike most previous studies that evaluate only on indoor scenes, we also tested the framework on challenging outdoor scenes \cite{knapitsch2017tanks} to demonstrate its generalization capability. Source code for our method is publicly available\footnote{https://github.com/lbg030/MVS-GS}
    
\end{itemize}

\section{RELATED WORKS}


\begin{figure*}[t]  
    \centering
    \includegraphics[width=0.84\textwidth]{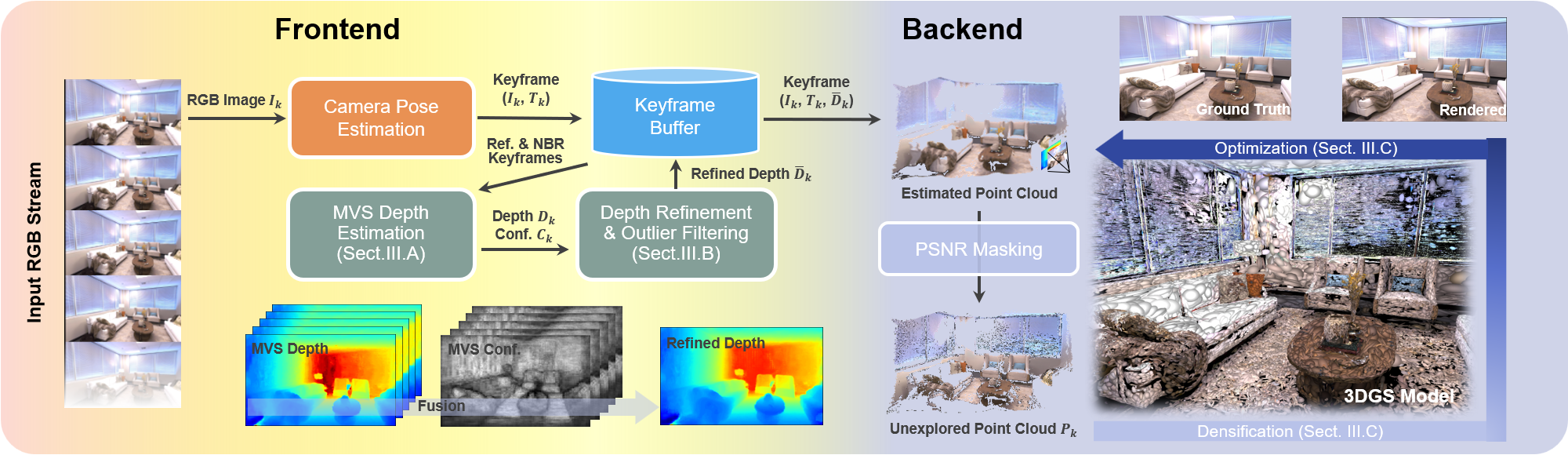}  
    \caption{System overview: Our system consists of a frontend and a backend, both running in parallel. The frontend initially estimates the camera pose of keyframes using SLAM. It then estimates the depth map and confidence map of each keyframe based on MVS, refining the depth map by incorporating depth information from sequential keyframes. The backend generates new Gaussian points from the refined depth map and integrates them into the 3DGS model. The backend then continuously optimizes the 3DGS model.}
    \label{fig2: System_framework}
    \vspace{-12pt}
\end{figure*}

\noindent \textbf{3D Modeling via Multi-View Stereo.} MVS reconstructs 3D models of scenes by identifying dense correspondences across multiple images \cite{schonberger2016pixelwise}. Many studies have addressed challenging issues such as matching ambiguity and high computational complexity. Learning-based MVS methods \cite{gu2020cascade} \cite{cao2022mvsformer} have generally proven more effective than traditional approaches \cite{schonberger2016pixelwise} in overcoming these challenges. Efforts have focused on improving performance by incorporating cascade cost volumes \cite{gu2020cascade}, or feature matching networks \cite{giang2021curvature} \cite{cao2022mvsformer}.

NeRF \cite{mildenhall2021nerf} models a scene as a radiance field and optimizes it through differentiable volume rendering, resulting in photorealistic view synthesis. However, this process requires extensive per-scene optimization. To overcome this limitation, some studies \cite{chen2021mvsnerf} \cite{liu2024geometry} have incorporated MVS techniques to generalize NeRF to unseen scenes. MVSNeRF \cite{chen2021mvsnerf} employs an explicit geometry-aware cost volume derived from MVS to model geometry in novel views.

Even though NeRF-based methods have achieved significant results, their performance is still limited by slow optimization and rendering speeds. As an alternative to NeRF, 3DGS \cite{kerbl20233d} has recently been proposed and has gained considerable popularity in neural rendering. 3DGS explicitly represents scenes using anisotropic 3D Gaussians, allowing for real-time, high-quality rendering through a differentiable tile-based rasterizer. Similar to NeRF, several studies \cite{zheng2024gps} \cite{liu2024fast} have attempted to apply MVS techniques to 3DGS to achieve generalized performance. GPS-Gaussian \cite{zheng2024gps} combines iterative stereo-matching-based depth estimation with pixel-wise Gaussian parameter regression. MVSGaussian \cite{liu2024fast} also employs MVS to estimate depth and establishes a pixel-aligned Gaussian representation. In \cite{zheng2024gps} \cite{liu2024fast}, the rich 3D information provided by MVS significantly enhances the rendering performance of the generated 3DGS. However, all of these methods are designed as offline processes, making real-time processing impossible.

\noindent \textbf{Monocular Dense SLAM.} Traditional SLAM research \cite{mur2015orb} \cite{qin2018vins} has primarily focused on accurate camera localization, yielding impressive results. Recently, many studies have shifted their focus toward precise 3D mapping, leading to the increased popularity of dense SLAM. Classic methods used motion stereo with handcrafted features for dense \cite{pizzoli2014remode} or semi-dense mapping \cite{engel2014lsd}.

Many studies have integrated deep learning models into monocular dense SLAM. DeepFactors \cite{czarnowski2020deepfactors} compresses dense depth maps into a low-dimensional latent space using deep learning techniques, thereby reducing the computational burden of depth estimation.
DROID-SLAM \cite{teed2021droid} employs an optical flow network to establish dense pixel correspondences and performs dense bundle adjustments. This approach achieves excellent trajectory estimation while simultaneously generating dense 3D models.

As NeRF \cite{mildenhall2021nerf} has advanced, many studies \cite{chung2023orbeez} \cite{matsuki2023imode} \cite{rosinol2023nerf} \cite{zhang2023go} \cite{zhu2024nicer} have explored its potential for representing 3D spaces in rendering and novel view synthesis. Orbeez-SLAM \cite{chung2023orbeez} and iMODE \cite{matsuki2023imode} directly integrate the sparse features or semi-dense depths obtained from ORB-SLAM \cite{mur2015orb} into a NeRF model. Similarly, NeRF-SLAM \cite{rosinol2023nerf} and GO-SLAM \cite{zhang2023go} utilize low-resolution depth maps estimated by DROID-SLAM for NeRF-based mapping. While some research \cite{zhu2024nicer} \cite{zhang2024glorie} has explored using depth prediction networks \cite{eftekhar2021omnidata} for mapping, the accuracy of these predicted depths is often limited by the training dataset and does not match the precision of MVS-based depth estimates.

Recently, 3DGS has shown great promise in 3D modeling by addressing the limitations of NeRF and offering faster rendering speeds. Although only a few studies \cite{huang2024photo} \cite{matsuki2024gaussian} have applied 3DGS to dense SLAM, these have demonstrated notably superior performance compared to NeRF-based approaches. Photo-SLAM \cite{huang2024photo} generates Gaussian points from the sparse features extracted by ORB-SLAM and then densifies them using geometric information through a Gaussian-Pyramid-based learning approach. MonoGS \cite{matsuki2024gaussian} estimates camera poses directly from images rendered by 3DGS, bypassing explicit depth estimation and instead generating new Gaussians randomly from the rendered depths.

These methods \cite{chung2023orbeez} \cite{matsuki2023imode} \cite{zhu2024nicer} \cite{huang2024photo} \cite{matsuki2024gaussian} rely on sparse or inaccurate depth data to construct NeRF or 3DGS models, resulting in low-quality outcomes. Most approaches directly adopt RGBD-based mapping techniques \cite{zhu2022nice} \cite{keetha2024splatam}, which depend on reliable depth maps. However, the depths derived from RGB images are often imprecise and noisy, ultimately reducing performance. In contrast, our method utilizes MVS to estimate high-quality depths and applies stringent noise filtering techniques to generate Gaussian points. As a result, this approach produces a 3DGS model that delivers more accurate rendering compared to existing methods.

\section{PROPOSED METHOD}

Our goal is to reconstruct a 3DGS model online for high-quality rendering from an RGB image stream. Fig. 2 describes the overall architecture of the proposed modeling system. The system is mainly composed of two modules: the \textit{frontend} and the \textit{backend}, which operate in parallel as independent threads.

The frontend first tracks camera poses $\left\{T_{k}\right\}$ of sequential input images $\left\{I_{k}\right\}$ by using the DROID-SLAM \cite{teed2021droid}. DROID-SLAM utilizes an optical flow network to predict dense pixel correspondences between adjacent frames. It then continuously performs dense bundle adjustment between keyframes to refine the camera poses. When a new keyframe is identified, the system performs MVS depth estimation by using the frame as a reference image (detailed in Section III.A). For each keyframe $F_k$ it saves the image-pose pair $(I_k, T_k)$ along with the estimated depth map $D_k$ and confidence map $C_k$ in a keyframe buffer. The estimated depth maps are relatively inaccurate and contain many outliers compared to offline MVS. Therefore, the system refines the depths and robustly filters out outliers based on previously estimated depth information (detailed in Section III.B).

The backend sequentially integrates the estimated depth maps into the 3DGS model (as detailed in Section III.C). It identifies unexplored regions in the depth map and converts only those depths into a point cloud $P_k$. This point cloud is then used to initialize new Gaussians $\left\{g_{i}\right\}$, which are subsequently added to the 3DGS model. The backend continuously optimizes the Gaussian parameters, running this process in parallel with the frontend while continuously appending new Gaussians from unexplored regions.

\subsection{Online MVS Depth Estimation}

We utilize the deep learning-based MVS method, MVSFormer \cite{cao2022mvsformer}, for estimating depth maps for each keyframe. MVSFormer uses a visual transformer to learn robust feature representations, achieving state-of-the-art results in MVS reconstruction. It features a cascade structure \cite{gu2020cascade} that utilizes multiple smaller cost volumes instead of a single large one, which significantly reduces GPU memory usage and runtime. This cascading strategy enables the real-time generation of high-resolution depth maps, making it well-suited for our online modeling system.

Given a keyframe $F_k$, we use $F_k$ as a reference frame and its neighboring keyframes $\left\{F^{nbr}_{k,n}\right\}$ as source frames to estimate the depth map $D_k$ and confidence map $C_k$. $\left\{F^{nbr}_{k,n}\right\}$ are determined as the $N_{nbr}$ consecutive keyframes before and after $F_k$. MVSFormer initially extracts multi-scale visual features from hierarchical vision transformers. These extracted features are used to construct cascade cost volumes at progressively finer scales using 3D CNNs \cite{gu2020cascade}. The depths estimated by DROID-SLAM are used to determine the depth hypothesis range for the first-stage cost volume. It then incrementally refines the depth maps from coarse to fine by narrowing the depth range and reducing the number of hypotheses. Finally, MVSFormer predicts a depth map $D_k$ and a confidence map $C_k$ from a cost-volume by using the temperature-based method \cite{cao2022mvsformer}. The method unifies the advantages of both regression-based depth and classification-based depth. The regression-based method (i.e., the expectation of depth probability) yields precise depth results. Conversely, using cross-entropy loss to optimize the network for depth classification offers a more reliable confidence estimation. By combining both methods, MVSFormer produces smooth and accurate depths and confidences.

\begin{figure}
    \centering
    \includegraphics[width=1\linewidth]{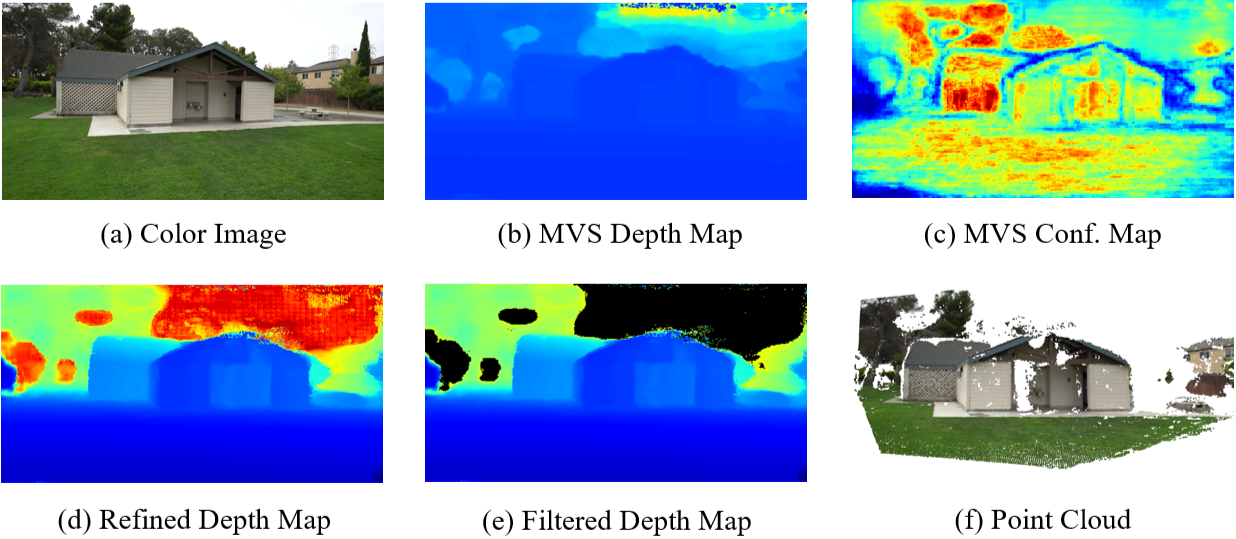}
    \caption{An illustration of the depth map refinement and filtering process: (b) the depth map and (c) the confidence map estimated by MVS are (d) refined using V-Fuse \cite{burgdorfer2023v}. Our method then (e) filters outliers by checking the geometric consistency of the depths.}
    \label{fig3: depth map refinement}
    \vspace{-13pt}
\end{figure}

\subsection{Depth Refinement and Outlier Filtering}

The estimated depth maps often contain numerous outliers, especially in the background and boundary regions. Additionally, online depth estimation tends to produce more outliers compared to offline methods due to limited source views. The source views can lead to incomplete coverage of the reference view and result in stereo matching errors from invisible pixels. To address these issues, our system incorporates additional steps to refine the depth maps and robustly filter outliers by leveraging sequentially estimated depth and confidence information.

Given the depth and confidence maps $\left(D_k,C_k\right)$ of the reference keyframe $F_k$, as well as the set of depth and confidence maps $\{(D^{nbr}_{k,n},C^{nbr}_{k,n})\}$ of the neighboring keyframes $\{F^{nbr}_{k,n}\}$, we integrate them to generate a refined depth and confidence maps $\left(\bar{D}_k,\bar{C}_k\right)$ for $F_k$. To solve this problem, we employ the depth map fusion network, \textit{V-Fuse} \cite{burgdorfer2023v}. It first calculates the mean and standard deviation of the input depth and confidence values to establish the depth hypothesis range for each pixel. It then constructs a visibility constraint volume by analyzing visibility factors such as support, occlusions, and free-space violations. This volume is regularized using a 3D convolutional network, after which the system generates the fused depth map through regression.

Next, we remove the outliers from the refined depth map. We first filter out low-confidence depths on $\bar{D}_k$ by applying a threshold to the confidence map $\bar{C}_k$. Then, it assesses the geometric consistency \cite{song2021view} of the depth $\bar{D}_k$ against the depths $\{D^{nbr}_{k,n}\}$ of $\{F^{nbr}_{k,n}\}$. For each neighboring keyframe $F^{nbr}_{k,n}$, the system wraps depths of $\bar{D}_k$ onto the view of $F^{nbr}_{k,n}$. It calculates the relative depth differences between the wrapped depths and the depths of $F^{nbr}_{k,n}$. If the relative depth difference is smaller than the defined threshold, the depth is considered geometrically consistent. Finally, depths that do not exhibit consistency across at least three views are filtered out. Fig. 3 illustrates the depth filtering process.

\subsection{Online 3DGS Mapping}
The backend module integrates the filtered depth maps $\bar{D}_k$ into a unified 3DGS model. To enhance efficiency, we employ the \textit{generalized exponential splatting} (GES) \cite{hamdi2024ges} in place of the original approach \cite{kerbl20233d} for constructing the 3DGS model. We have adapted the GES method to support online optimization of the 3DGS by sequentially inputting keyframes. The module continuously produces new Gaussian points from the incoming keyframes and updates their parameters through iterative processes. This operation is carried out in parallel with the frontend module.

\noindent \textbf{Generalized Exponential Splatting}.
The 3DGS model represents the scene using a set of 3D Gaussians, $\mathcal{G}=\{g_i\}$, where each Gaussian point $g_i$ is defined by a covariance matrix $\Sigma_i \in R^{3 \times 3}$, a mean $\mu_i \in R^3$, opacity $o_i \in [0,1]$, and color $c_i \in R^3$. Original 3DGS requires a large number of points to capture high-frequency details, making it inefficient for real-time mapping and rendering, particularly in large-scale scenes. To address this, we employ the more efficient method, GES, which is an extension of the generalized exponential function (GEF). Density function for each Gaussian with respect to a 3D point $x$ is given by the GEF as:
\[
g_i(x) = \exp\left\{ -\frac{1}{2} (x - \mu_i)^T \Sigma_i^{-1} (x - \mu_i) \right\}^{\frac{\beta}{2}}
\]
where $\beta$ is a shape parameter that controls the splat's sharpness. GES treats this shape parameter as a trainable element across different frequency bands, leading to more adaptable and efficient rendering. This method effectively handles high-frequency components that posed challenges for the original 3DGS approach. GES also incorporates edge recognition via an edge-aware mask, which is generated using a difference of Gaussian filter.

To render an image $\hat{I}_k$ from an input pose $\hat{T}_k$, the 3D Gaussians are projected onto the image plane. The color $\hat{c}$ of each pixel $p$ is determined by sorting the Gaussian points by depth and applying front-to-back alpha-blending, as follows:
\[
\hat{c}(p) = \sum_{i \in N} c_i \alpha_i \prod_{j=1}^{i-1} (1 - \alpha_j)
\]
where $\alpha_i$ is calculated as the product of the opacity $o_i$ and the GEF density $g_i$ \cite{hamdi2024ges}. This method ensures that the closer Gaussian points contribute more prominently to the final color, while the further points blend into the background.

The loss function used to train the GES is defined as \cite{hamdi2024ges}:
\[
\mathcal{L}=\lambda_{L1}\mathcal{L}_1+\lambda_{ssim}\mathcal{L}_{ssim}+\lambda_\omega\mathcal{L}_\omega
\]
where $L_{ssim}$ refers to the structural similarity loss. $L_\omega$ represents the frequency-modulated loss derived from the edge-aware mask. The $L_\omega$ term ensures that the shape parameter is trained over a wide frequency range, allowing images to be rendered with fewer splats while enhancing edge sharpness.

\noindent \textbf{Online Densification}. The backend module incrementally densifies the 3DGS model by initializing new Gaussians at each incoming keyframe. We transform the filtered depth map $\bar{D}_k$ of a keyframe $F_k$ into a 3D point cloud $P_k$, which is then used for densification. Since $P_k$ contains a large number of 3D points, it would be inefficient to directly append all of them as new Gaussians. Therefore, we segment the depth map $\bar{D}_k$ into explored and unexplored regions, creating the point cloud $P_k$ only from the depths in the unexplored regions. The explored region consists of areas where the current Gaussians already accurately represent the scene geometry, while the unexplored region includes areas where the Gaussians are either insufficient or newly observed. A pixel is classified as part of the unexplored region if its rendered color significantly differs from the original color. To identify such pixels, we compute the PSNR between the original and rendered images, marking those with a PSNR below a certain threshold as belonging to the unexplored region.

The extracted point cloud $P_k$ might contain multiple points that are densely clustered in certain areas. To address this, we further reduce the number of points in $P_k$ using a voxel grid filter. The filtered point cloud is then directly converted into new Gaussian points $\{g_i^{new}\}$, with each point assigned an identity covariance matrix  $\Sigma_i$ and $\beta = 2$. Finally, these new Gaussian points are added to the 3DGS $\mathcal{G}$. 

\noindent \textbf{Optimization}. The backend module iteratively updates the parameters of the 3DGS model using differentiable rendering and gradient-based optimization \cite{kerbl20233d}. It also performs adaptive density control to optimize the distribution of Gaussian points. This method dynamically adjusts the point density according to the level of detail required in different regions of the scene, ensuring efficient use of computational resources by assigning more Gaussians to high-frequency detail areas and fewer to low-frequency regions.

In each iteration, we use the current keyframe along with previously input keyframes for optimization. Due to limited computational time, instead of using all input keyframes, we randomly select a subset of keyframes.
The module performs optimization with 1,000 iterations every time a new frame is input: 500 using frames from a specific time window around the current frame, and 500 from the entire set.
This approach helps mitigate the forgetting problem, where the influence of earlier input frames gradually diminishes over time.

\begin{table}[t]
\caption{Quantitative evaluation on the Replica RGB dataset}
\centering
\scalebox{0.70}{
\begin{tabular}{lccccccccc}
\midrule
Method & Off0 & Off1 & Off2 & Off3 & Off4 & Rm0 & Rm1 & Rm2 & Avg. \\
\midrule
\multicolumn{10}{c}{\textbf{PSNR} $\uparrow$} \\ 
\midrule
Go-SLAM      & -     & -     & -     & -     & -     & -     & -     & -     & 21.17 \\
NICER-SLAM   & 28.54 & 25.86 & 21.95 & 26.13 & 25.47 & 25.33 & 23.92 & 26.12 & 25.41 \\
GLORIE-SLAM  & 35.88 & 37.15 & 28.45 & 28.54 & 29.73 & 28.49 & 30.09 & 29.98 & 31.04 \\
Q-SLAM       & 36.31 & 37.22 & 30.68 & 30.21 & 31.96 & 29.58 & \textbf{32.74} & 31.25 & 32.49 \\
Photo-SLAM   & \underline{36.99} & \underline{37.52} & \underline{31.79} & \underline{31.62} & \textbf{34.17} & \underline{29.77} & 31.30 & \underline{33.18} & \underline{33.29} \\
MonoGS       & 32.00 & 31.21 & 23.26 & 25.77 & 23.85 & 23.53 & 25.00 & 22.42 & 25.88 \\
Ours         & \textbf{41.02} & \textbf{42.04} & \textbf{34.00} & \textbf{34.65} & \underline{33.33} & \textbf{32.20} & \underline{31.54} & \textbf{35.84} & \textbf{35.58} \\
\midrule
\multicolumn{10}{c}{\textbf{SSIM} $\uparrow$} \\ 
\midrule
Go-SLAM      & -     & -     & -     & -     & -     & -     & -     & -     & 0.70 \\
NICER-SLAM   & 0.87  & 0.85  & 0.82  & 0.86  & 0.87  & 0.75  & 0.77  & 0.83  & 0.83 \\
GLORIE-SLAM  & \underline{0.97}  & \textbf{0.99}  & \textbf{0.97}  & \textbf{0.97}  & \textbf{0.97}  & \textbf{0.96}  & \textbf{0.97}  & \textbf{0.96}  & \textbf{0.97} \\
Q-SLAM       & 0.94  & 0.94  & 0.90  & 0.88  & 0.89  & 0.83  & 0.91  & 0.87  & 0.89 \\
Photo-SLAM   & 0.96  & 0.95  & 0.93  & 0.92  & 0.94  & 0.87  & 0.91  & \underline{0.93}  & 0.93 \\
MonoGS       & 0.90  & 0.88  & 0.82  & 0.84  & 0.86  & 0.75  & 0.79  & 0.81  & 0.83 \\
Ours         & \textbf{0.98}  & \underline{0.98}  & \underline{0.95}  & \underline{0.96}  & \underline{0.95}  & \underline{0.95}  & \underline{0.92}  & \textbf{0.96}  & \underline{0.96} \\
\midrule
\multicolumn{10}{c}{\textbf{LPIPS} $\downarrow$} \\ 
\midrule
Go-SLAM      & -     & -     & -     & -     & -     & -     & -     & -     & 0.41 \\
NICER-SLAM   & 0.17  & 0.18  & 0.20  & 0.16  & 0.18  & 0.25  & 0.22  & 0.18  & 0.19 \\
GLORIE-SLAM  & 0.09  & 0.08  & \underline{0.15}  & 0.11  & 0.15  & \underline{0.13}  & \underline{0.13}  & \underline{0.14}  & \underline{0.12} \\
Q-SLAM       & 0.13  & 0.15  & 0.20  & 0.19  & 0.18  & 0.18  & 0.16  & 0.15  & 0.17 \\
Photo-SLAM   & \underline{0.06}  & \underline{0.06}  & \textbf{0.09}  & \underline{0.09}  & \textbf{0.07}  & \textbf{0.10}  & \textbf{0.08}  & \textbf{0.07}  & \textbf{0.08} \\
MonoGS       & 0.23  & 0.22  & 0.30  & 0.24  & 0.34  & 0.33  & 0.35  & 0.39  & 0.30 \\
Ours         & \textbf{0.05}  & \textbf{0.05}  & \textbf{0.09}  & \textbf{0.07}  & \underline{0.10}  & \textbf{0.10}  & \underline{0.13}  & \textbf{0.07}  & \textbf{0.08} \\
\bottomrule
\vspace{-15pt}
\end{tabular}
}
\end{table}


\begin{table}[t]
\caption{Quantitative evaluation on the TUM-RGBD dataset}
\centering
\scalebox{1}{
\begin{tabular}{llcccc}
\toprule
Metrics & Method & f1/desk & f2/xyz & f3/off & Avg. \\
\midrule
\multirow{5}{*}{PSNR $\uparrow$}  & Go-SLAM     & 11.71   &14.81   &13.57   &13.36  \\
                                  & MonoGS      & 19.67   &16.17   &20.63   &18.82  \\
                                  & Photo-SLAM  & \textbf{20.97}   &21.07   &19.59   &20.54  \\
                                  & GLORIE-SLAM & 20.26   &\textbf{25.62}   &\underline{21.21}   &\underline{22.36}  \\
                                  & Ours        & \underline{20.67}   &\underline{24.53}   &\textbf{22.37}   &\textbf{22.52}  \\
\midrule
\multirow{5}{*}{SSIM $\uparrow$}  & Go-SLAM     & 0.41    & 0.44   & 0.48   & 0.44 \\
                                  & MonoGS      & 0.73    & 0.72   & 0.77   & 0.74 \\
                                  & Photo-SLAM  & 0.74    & 0.73   & 0.69   & 0.72 \\
                                  & GLORIE-SLAM & \textbf{0.87}    & \textbf{0.96}   & \textbf{0.84}   & \textbf{0.89} \\
                                  & Ours        & \underline{0.77}    & \underline{0.86}   & \underline{0.80}   & \underline{0.81} \\
\midrule
\multirow{5}{*}{LPIPS $\downarrow$} & GO-SLAM   & 0.61    & 0.57   & 0.64   & 0.61 \\
                                  & MonoGS      & 0.33    & 0.31   & 0.34   & 0.33 \\
                                  & Photo-SLAM  & \textbf{0.23}    & 0.17   & \textbf{0.24}   & \textbf{0.21} \\
                                  & GLORIE-SLAM & 0.31    & \textbf{0.09}   & \underline{0.32}   & \underline{0.24} \\
                                  & Ours        & \underline{0.25}    & \underline{0.15}   & \textbf{0.24}   & \textbf{0.21} \\
\bottomrule
\end{tabular}
}
\end{table}

\begin{figure}
    \centering
    \includegraphics[width=0.95\linewidth]{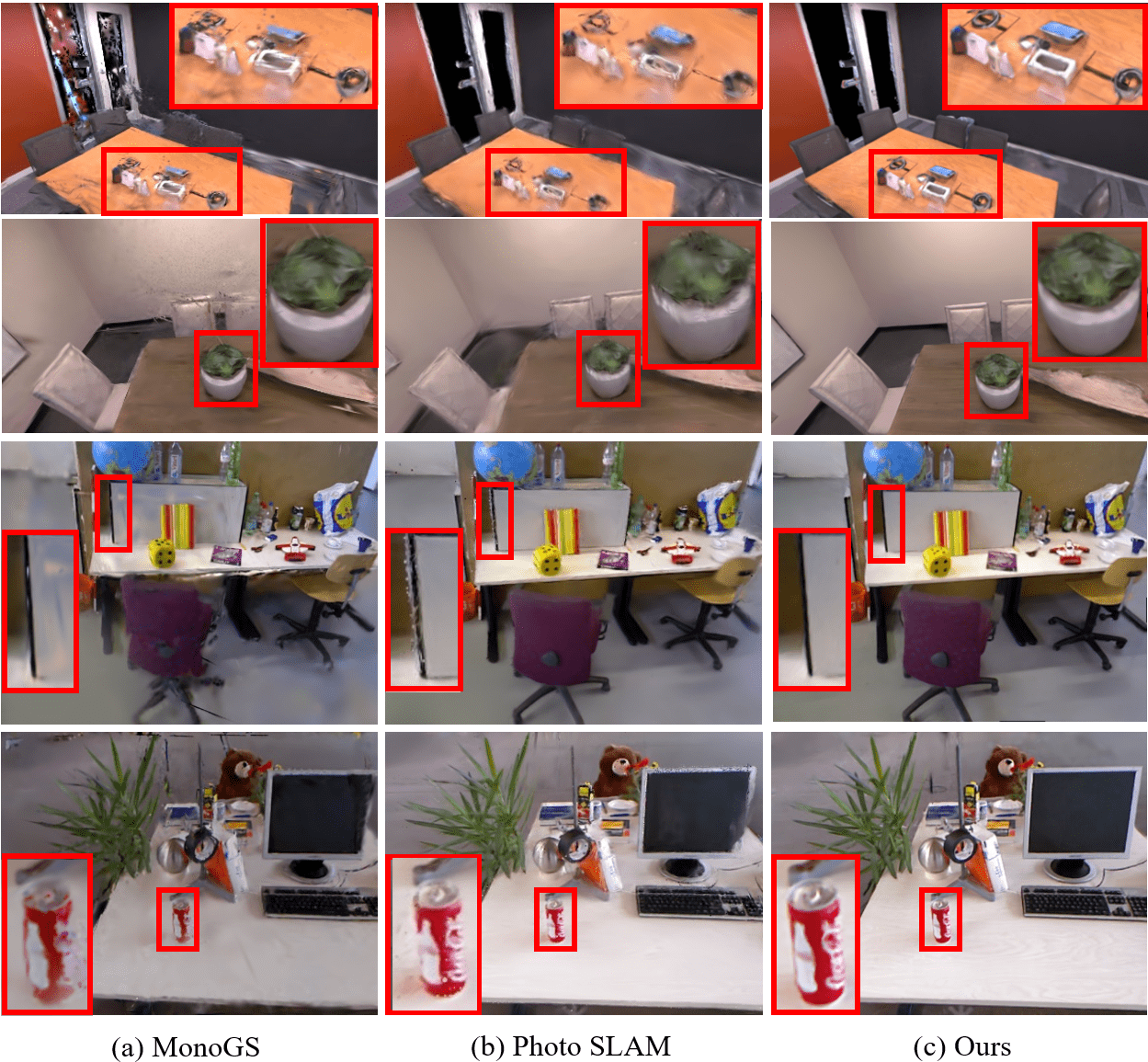}
    \caption{Qualitative evaluation on the Replica RGB dataset (first and second rows) and the TUM-RGBD dataset (third and fourth rows).}
    \label{fig4: Qualitative evaluation on Replica}
    \vspace{-10pt}
\end{figure}

\section{EXPERIMENTAL RESULTS}

To verify the performance of our method, we conducted comparative experiments on online 3D modeling for neural rendering. We evaluated the modeling performance on both indoor and outdoor scenes. Additionally, we performed an ablation study to assess the effectiveness of key components. 

All experiments were performed on a desktop computer featuring an Intel i9-13900KS processor and an NVIDIA GeForce RTX 3090 Ti GPU.
CUDA was employed for GPU acceleration in some processes, including time-sensitive rasterization and gradient computation, while the remaining components were implemented using PyTorch.

\subsection{Evaluation in Indoor Scenes}
We evaluated the rendering performance of our method in indoor scenes using the Replica dataset \cite{straub2019replica} (8 sequences) and the TUM RGBD dataset \cite{sturm2012benchmark} (3 sequences). Although these datasets include RGBD frames, we focused solely on the RGB images. For MVS depth estimation, we used $384 \times 512$ images, while 3DGS was trained on the original resolution. We set the voxel resolution to $0.005m$, confidence threshold to $0.7$, the number of neighboring keyframes $N_{nbr}$ to $4$, and PSNR threshold for masking to $40$.

The performance of our method was compared against state-of-the-art monocular dense SLAM methods, including NeRF-based methods (Go-SLAM \cite{zhang2023go}, NICER-SLAM \cite{rosinol2023nerf}, GLORIE-SLAM \cite{zhang2024glorie}, and Q-SLAM \cite{peng2024q}) and 3DGS-based methods (Photo-SLAM \cite{huang2024photo} and MonoGS \cite{matsuki2024gaussian}). For evaluating rendering performance, we utilized standard photometric metrics: PSNR, SSIM \cite{wang2004image}, and LPIPS \cite{zhang2018unreasonable}.

Tables I and II show the rendering accuracy on the Replica and TUM-RGBD datasets, respectively. As illustrated in the tables, our method achieves the highest performance in average PSNR and LPIPS, while also securing either the top or second-best results in average SSIM. Notably, our method significantly outperforms MonoGS. MonoGS generates Gaussian points based on arbitrary depth estimates for unexplored scenes, which leads to inaccurate depth information and consequently produces low-quality 3DGS maps with poor rendering results. Additionally, our method surpasses Photo-SLAM in all metrics, demonstrating that generating Gaussian points from dense MVS depths is much more effective in producing high-quality 3DGS models compared to the sparse feature point-based initialization method employed by Photo-SLAM.

Fig. 4 shows the qualitative comparison results of 3DGS-based methods. As shown in Fig. 4, our method produces high-quality rendering results with greater scene detail and fewer artifacts compared to the other methods.

\begin{figure}[t]
    \centering
    \includegraphics[width=0.95\linewidth]{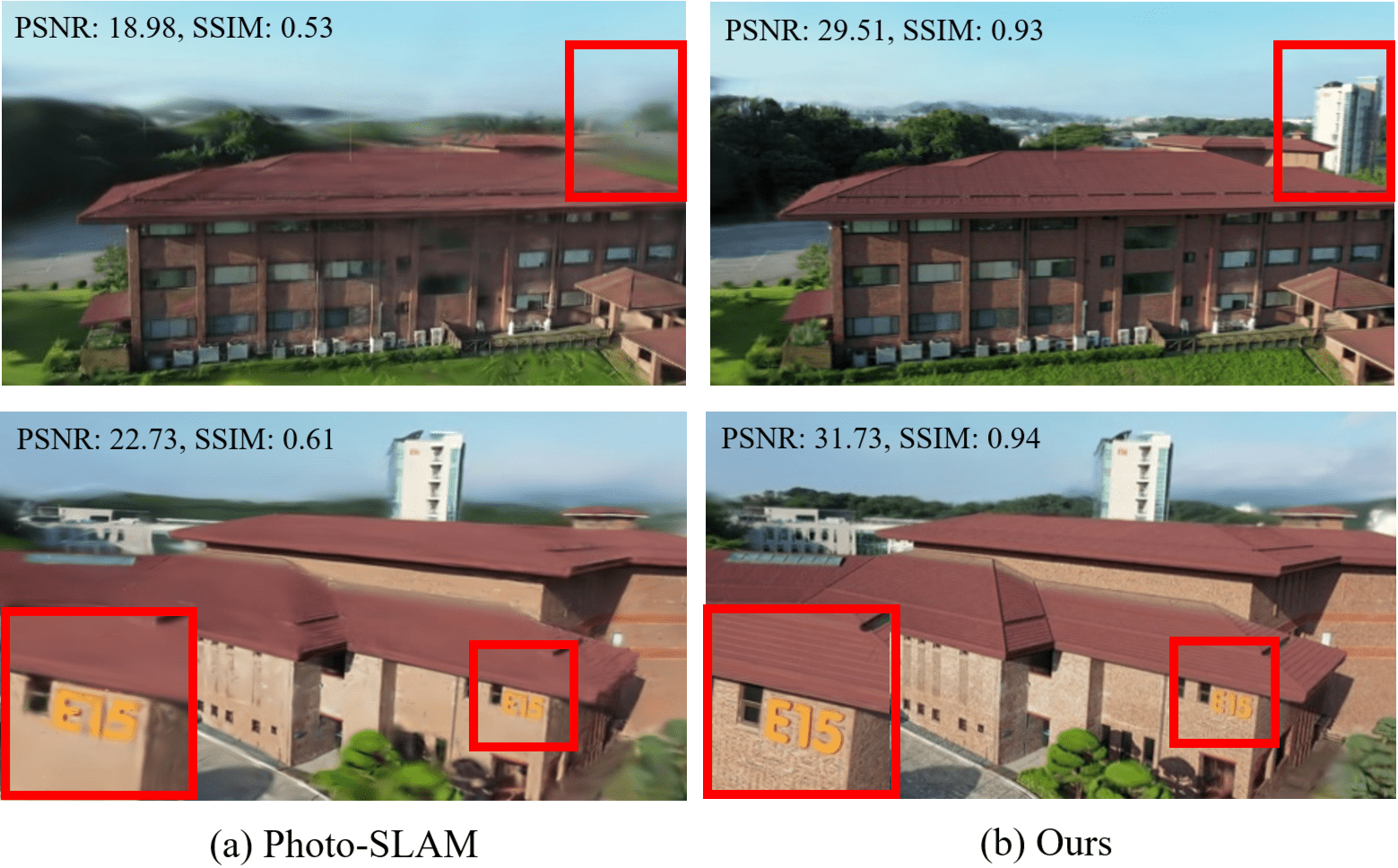}
    \caption{Rendering results of (a) Photo-SLAM \cite{huang2024photo} and (b) our method on two aerial scenes. Each PSNR and SSIM represents the average result for the entire scene.}
    \label{fig5: rendering results}
    \vspace{-12pt}
\end{figure}

\subsection{Evaluation in Outdoor Scenes}
This section assesses the generalization capability of our method by evaluating its performance on outdoor scenes, including aerial scenes \cite{song2021view} and the Tanks and Temples dataset \cite{knapitsch2017tanks}. Fig. 5 shows the rendering results along with the PSNR and SSIM performance of the generated 3DGS in two aerial scenes.
Both Photo-SLAM and our method successfully generated the 3DGS in the scenes, but MonoGS failed to produce the 3DGS model.
As illustrated in Fig. 5, our method outperformed Photo-SLAM in terms of PSNR and SSIM, leading to more accurate rendering results. The disparity in performance between Photo-SLAM and our method is considerably more pronounced in outdoor scenes than in indoor scenes. Since MVS methods can accurately reconstruct large-scale scenes over wide ranges, our method proves to be significantly more effective in generating 3DGS models in aerial scenes.

Fig. 6 shows the rendering performance of our method in several outdoor scenes from the Tanks and Temples dataset. In these scenes, both Photo-SLAM and MonoGS failed to generate 3DGS models. As can be seen in Fig. 6, our method achieved highly accurate rendering results, closely resembling real pictures.

\begin{figure}[t]
    \centering
    \includegraphics[width=0.95\linewidth]{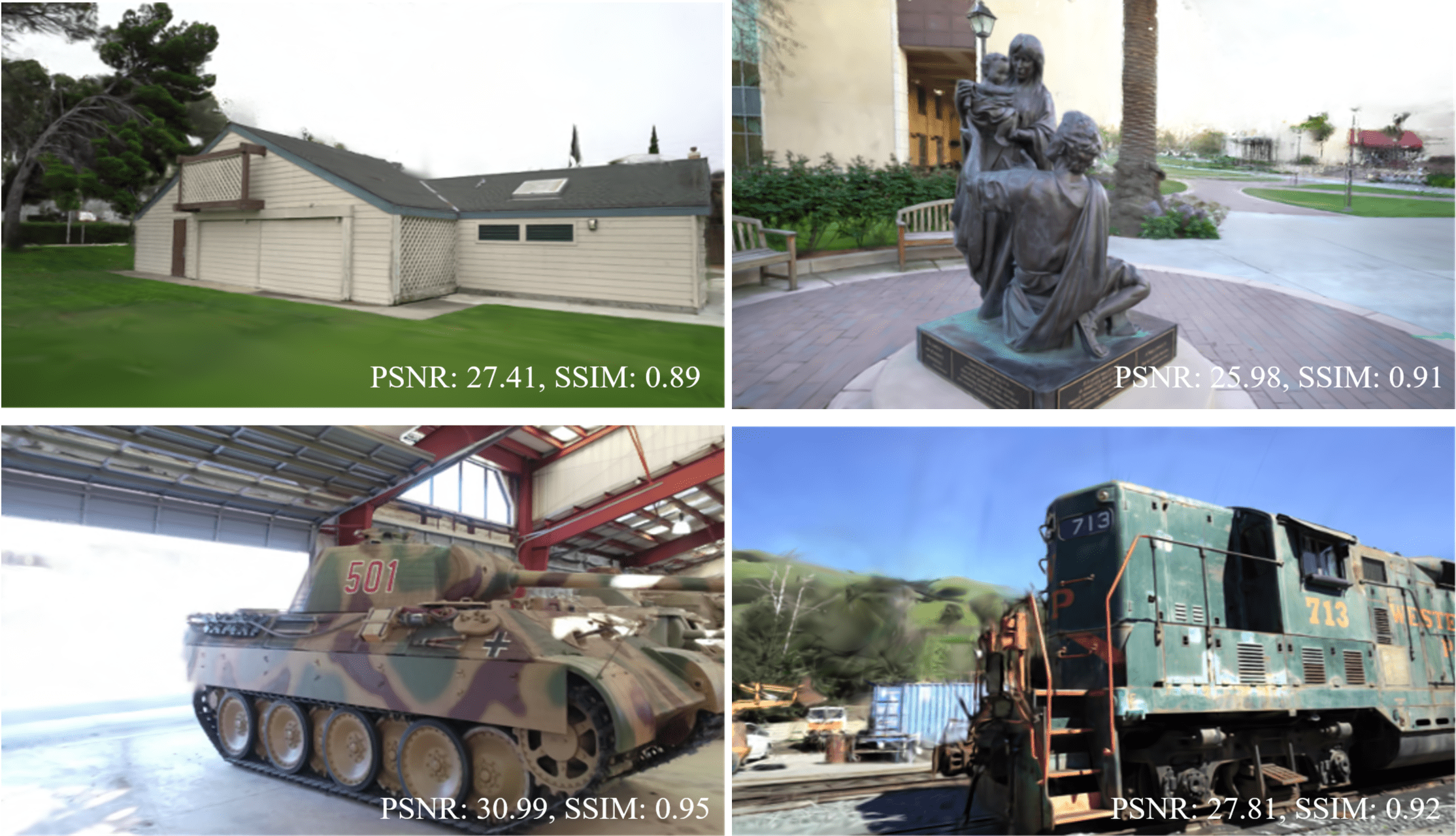}
    \caption{Rendering results of our method on the Tanks and Temples dataset.}
    \label{fig6: rendering results}
\end{figure}

\subsection{Ablation Study}
We performed an ablation study on the Office0 scene from the Replica dataset to validate the effectiveness of the key components of our method. We measured the photometric metrics and the number of keyframes processed per second (FPS). Table III shows the performance variation of each key component (MVS depth estimation in Sect. III.A, depth filtering in Sect. III.B, and mask-based initialization of Gaussians in Sect. III.C) by comparing the four variants.

Method-A generates the 3DGS model directly from the low-resolution depth map of DROID-SLAM, resulting in high FPS but the lowest rendering performance. Method-C delivered the best rendering performance but had the lowest FPS. Interestingly, while Method-D slightly reduced rendering performance compared to Method-C, it achieved a higher FPS. This indicates that initializing Gaussian points only in the unexplored regions is an effective strategy.

\begin{table}[t]
\caption{Ablation Study on the Replica (Office0)}
\centering
\scalebox{0.8}{
\begin{tabular}{cccccccc}
\toprule
\multirow{2}{*}{\centering \textbf{Method}} & \textbf{MVS} & \textbf{Noise} & \textbf{Mask} & \multirow{2}{*}{\centering \textbf{PSNR}} & \multirow{2}{*}{\centering \textbf{SSIM}} & \multirow{2}{*}{\centering \textbf{LPIPS}} & \multirow{2}{*}{\centering \textbf{FPS}} \\
                & \textbf{Depth} & \textbf{Filtering} & \textbf{Update} &  &  &  &  \\
\midrule
Method-A &  X         & X               & X           & 23.46  & 0.84  & 0.54  & 18.96 \\
Method-B &  O         & X               & X           & 40.74  & 0.97  & 0.05  & 13.72 \\
Method-C &  O         & O               & X           & 41.30  & 0.97  & 0.04  & 10.02 \\
Method-D &  O         & O               & O           & 41.02  & 0.97  & 0.05  & 11.56 \\
\bottomrule
\vspace{-15pt}
\end{tabular}}
\end{table}

\section{CONCLUSION}
We present a new online mapping method for high-quality 3DGS modeling. This method uses MVS to generate high-resolution depth maps and initializes Gaussian points accordingly. By integrating sequential depth information, we achieve highly accurate and noise-filtered depth estimates. This results in the generation of highly precise 3DGS models. Our experiments demonstrate that our method outperforms state-of-the-art dense SLAM methods, with exceptional performance in outdoor scenes.

\bibliographystyle{IEEEtran}
\bibliography{IEEEabrv,root}

\begin{thebibliography}{10}
\providecommand{\url}[1]{#1}
\csname url@rmstyle\endcsname
\providecommand{\newblock}{\relax}
\providecommand{\bibinfo}[2]{#2}
\providecommand\BIBentrySTDinterwordspacing{\spaceskip=0pt\relax}
\providecommand\BIBentryALTinterwordstretchfactor{4}
\providecommand\BIBentryALTinterwordspacing{\spaceskip=\fontdimen2\font plus
\BIBentryALTinterwordstretchfactor\fontdimen3\font minus \fontdimen4\font\relax}
\providecommand\BIBforeignlanguage[2]{{%
\expandafter\ifx\csname l@#1\endcsname\relax
\typeout{** WARNING: IEEEtran.bst: No hyphenation pattern has been}%
\typeout{** loaded for the language `#1'. Using the pattern for}%
\typeout{** the default language instead.}%
\else
\language=\csname l@#1\endcsname
\fi
#2}}

\bibitem{schonberger2016pixelwise}
J.~L. Sch{\"o}nberger, E.~Zheng, J.-M. Frahm, and M.~Pollefeys, ``Pixelwise view selection for unstructured multi-view stereo,'' in \emph{Computer Vision--ECCV 2016: 14th European Conference, Amsterdam, The Netherlands, October 11-14, 2016, Proceedings, Part III 14}.\hskip 1em plus 0.5em minus 0.4em\relax Springer, 2016, pp. 501--518.

\bibitem{gu2020cascade}
X.~Gu, Z.~Fan, S.~Zhu, Z.~Dai, F.~Tan, and P.~Tan, ``Cascade cost volume for high-resolution multi-view stereo and stereo matching,'' in \emph{Proceedings of the IEEE/CVF conference on computer vision and pattern recognition}, 2020, pp. 2495--2504.

\bibitem{cao2022mvsformer}
C.~Cao, X.~Ren, and Y.~Fu, ``Mvsformer: Multi-view stereo by learning robust image features and temperature-based depth,'' \emph{arXiv preprint arXiv:2208.02541}, 2022.

\bibitem{mildenhall2021nerf}
B.~Mildenhall, P.~P. Srinivasan, M.~Tancik, J.~T. Barron, R.~Ramamoorthi, and R.~Ng, ``Nerf: Representing scenes as neural radiance fields for view synthesis,'' \emph{Communications of the ACM}, vol.~65, no.~1, pp. 99--106, 2021.

\bibitem{chen2021mvsnerf}
A.~Chen, Z.~Xu, F.~Zhao, X.~Zhang, F.~Xiang, J.~Yu, and H.~Su, ``Mvsnerf: Fast generalizable radiance field reconstruction from multi-view stereo,'' in \emph{Proceedings of the IEEE/CVF international conference on computer vision}, 2021, pp. 14\,124--14\,133.

\bibitem{liu2024geometry}
T.~Liu, X.~Ye, M.~Shi, Z.~Huang, Z.~Pan, Z.~Peng, and Z.~Cao, ``Geometry-aware reconstruction and fusion-refined rendering for generalizable neural radiance fields,'' in \emph{Proceedings of the IEEE/CVF Conference on Computer Vision and Pattern Recognition}, 2024, pp. 7654--7663.

\bibitem{kerbl20233d}
B.~Kerbl, G.~Kopanas, T.~Leimk{\"u}hler, and G.~Drettakis, ``3d gaussian splatting for real-time radiance field rendering.'' \emph{ACM Trans. Graph.}, vol.~42, no.~4, pp. 139--1, 2023.

\bibitem{zheng2024gps}
S.~Zheng, B.~Zhou, R.~Shao, B.~Liu, S.~Zhang, L.~Nie, and Y.~Liu, ``Gps-gaussian: Generalizable pixel-wise 3d gaussian splatting for real-time human novel view synthesis,'' in \emph{Proceedings of the IEEE/CVF Conference on Computer Vision and Pattern Recognition}, 2024, pp. 19\,680--19\,690.

\bibitem{liu2024fast}
T.~Liu, G.~Wang, S.~Hu, L.~Shen, X.~Ye, Y.~Zang, Z.~Cao, W.~Li, and Z.~Liu, ``Fast generalizable gaussian splatting reconstruction from multi-view stereo,'' \emph{arXiv preprint arXiv:2405.12218}, 2024.

\bibitem{newcombe2011dtam}
R.~A. Newcombe, S.~J. Lovegrove, and A.~J. Davison, ``Dtam: Dense tracking and mapping in real-time,'' in \emph{2011 international conference on computer vision}.\hskip 1em plus 0.5em minus 0.4em\relax IEEE, 2011, pp. 2320--2327.

\bibitem{pizzoli2014remode}
M.~Pizzoli, C.~Forster, and D.~Scaramuzza, ``Remode: Probabilistic, monocular dense reconstruction in real time,'' in \emph{2014 IEEE international conference on robotics and automation (ICRA)}.\hskip 1em plus 0.5em minus 0.4em\relax IEEE, 2014, pp. 2609--2616.

\bibitem{koestler2022tandem}
L.~Koestler, N.~Yang, N.~Zeller, and D.~Cremers, ``Tandem: Tracking and dense mapping in real-time using deep multi-view stereo,'' in \emph{Conference on Robot Learning}.\hskip 1em plus 0.5em minus 0.4em\relax PMLR, 2022, pp. 34--45.

\bibitem{chung2023orbeez}
C.-M. Chung, Y.-C. Tseng, Y.-C. Hsu, X.-Q. Shi, Y.-H. Hua, J.-F. Yeh, W.-C. Chen, Y.-T. Chen, and W.~H. Hsu, ``Orbeez-slam: A real-time monocular visual slam with orb features and nerf-realized mapping,'' in \emph{2023 IEEE International Conference on Robotics and Automation (ICRA)}.\hskip 1em plus 0.5em minus 0.4em\relax IEEE, 2023, pp. 9400--9406.

\bibitem{matsuki2023imode}
H.~Matsuki, E.~Sucar, T.~Laidow, K.~Wada, R.~Scona, and A.~J. Davison, ``imode: Real-time incremental monocular dense mapping using neural field,'' in \emph{2023 IEEE International Conference on Robotics and Automation (ICRA)}.\hskip 1em plus 0.5em minus 0.4em\relax IEEE, 2023, pp. 4171--4177.

\bibitem{rosinol2023nerf}
A.~Rosinol, J.~J. Leonard, and L.~Carlone, ``Nerf-slam: Real-time dense monocular slam with neural radiance fields,'' in \emph{2023 IEEE/RSJ International Conference on Intelligent Robots and Systems (IROS)}.\hskip 1em plus 0.5em minus 0.4em\relax IEEE, 2023, pp. 3437--3444.

\bibitem{zhang2023go}
Y.~Zhang, F.~Tosi, S.~Mattoccia, and M.~Poggi, ``Go-slam: Global optimization for consistent 3d instant reconstruction,'' in \emph{Proceedings of the IEEE/CVF International Conference on Computer Vision}, 2023, pp. 3727--3737.

\bibitem{zhu2024nicer}
Z.~Zhu, S.~Peng, V.~Larsson, Z.~Cui, M.~R. Oswald, A.~Geiger, and M.~Pollefeys, ``Nicer-slam: Neural implicit scene encoding for rgb slam,'' in \emph{2024 International Conference on 3D Vision (3DV)}.\hskip 1em plus 0.5em minus 0.4em\relax IEEE, 2024, pp. 42--52.

\bibitem{zhang2024glorie}
G.~Zhang, E.~Sandstr{\"o}m, Y.~Zhang, M.~Patel, L.~Van~Gool, and M.~R. Oswald, ``Glorie-slam: Globally optimized rgb-only implicit encoding point cloud slam,'' \emph{arXiv preprint arXiv:2403.19549}, 2024.

\bibitem{peng2024q}
C.~Peng, C.~Xu, Y.~Wang, M.~Ding, H.~Yang, M.~Tomizuka, K.~Keutzer, M.~Pavone, and W.~Zhan, ``Q-slam: Quadric representations for monocular slam,'' \emph{arXiv preprint arXiv:2403.08125}, 2024.

\bibitem{huang2024photo}
H.~Huang, L.~Li, H.~Cheng, and S.-K. Yeung, ``Photo-slam: Real-time simultaneous localization and photorealistic mapping for monocular stereo and rgb-d cameras,'' in \emph{Proceedings of the IEEE/CVF Conference on Computer Vision and Pattern Recognition}, 2024, pp. 21\,584--21\,593.

\bibitem{matsuki2024gaussian}
H.~Matsuki, R.~Murai, P.~H. Kelly, and A.~J. Davison, ``Gaussian splatting slam,'' in \emph{Proceedings of the IEEE/CVF Conference on Computer Vision and Pattern Recognition}, 2024, pp. 18\,039--18\,048.

\bibitem{lan2024monocular}
T.~Lan, Q.~Lin, and H.~Wang, ``Monocular gaussian slam with language extended loop closure,'' \emph{arXiv preprint arXiv:2405.13748}, 2024.

\bibitem{zhu2022nice}
Z.~Zhu, S.~Peng, V.~Larsson, W.~Xu, H.~Bao, Z.~Cui, M.~R. Oswald, and M.~Pollefeys, ``Nice-slam: Neural implicit scalable encoding for slam,'' in \emph{Proceedings of the IEEE/CVF conference on computer vision and pattern recognition}, 2022, pp. 12\,786--12\,796.

\bibitem{keetha2024splatam}
N.~Keetha, J.~Karhade, K.~M. Jatavallabhula, G.~Yang, S.~Scherer, D.~Ramanan, and J.~Luiten, ``Splatam: Splat track \& map 3d gaussians for dense rgb-d slam,'' in \emph{Proceedings of the IEEE/CVF Conference on Computer Vision and Pattern Recognition}, 2024, pp. 21\,357--21\,366.

\bibitem{straub2019replica}
J.~Straub, T.~Whelan, L.~Ma, Y.~Chen, E.~Wijmans, S.~Green, J.~J. Engel, R.~Mur-Artal, C.~Ren, S.~Verma, \emph{et~al.}, ``The replica dataset: A digital replica of indoor spaces,'' \emph{arXiv preprint arXiv:1906.05797}, 2019.

\bibitem{sturm2012benchmark}
J.~Sturm, N.~Engelhard, F.~Endres, W.~Burgard, and D.~Cremers, ``A benchmark for the evaluation of rgb-d slam systems,'' in \emph{2012 IEEE/RSJ international conference on intelligent robots and systems}.\hskip 1em plus 0.5em minus 0.4em\relax IEEE, 2012, pp. 573--580.

\bibitem{knapitsch2017tanks}
A.~Knapitsch, J.~Park, Q.-Y. Zhou, and V.~Koltun, ``Tanks and temples: Benchmarking large-scale scene reconstruction,'' \emph{ACM Transactions on Graphics (ToG)}, vol.~36, no.~4, pp. 1--13, 2017.

\bibitem{giang2021curvature}
K.~T. Giang, S.~Song, and S.~Jo, ``Curvature-guided dynamic scale networks for multi-view stereo,'' \emph{arXiv preprint arXiv:2112.05999}, 2021.

\bibitem{mur2015orb}
R.~Mur-Artal, J.~M.~M. Montiel, and J.~D. Tardos, ``Orb-slam: a versatile and accurate monocular slam system,'' \emph{IEEE transactions on robotics}, vol.~31, no.~5, pp. 1147--1163, 2015.

\bibitem{qin2018vins}
T.~Qin, P.~Li, and S.~Shen, ``Vins-mono: A robust and versatile monocular visual-inertial state estimator,'' \emph{IEEE transactions on robotics}, vol.~34, no.~4, pp. 1004--1020, 2018.

\bibitem{engel2014lsd}
J.~Engel, T.~Sch{\"o}ps, and D.~Cremers, ``Lsd-slam: Large-scale direct monocular slam,'' in \emph{European conference on computer vision}.\hskip 1em plus 0.5em minus 0.4em\relax Springer, 2014, pp. 834--849.

\bibitem{czarnowski2020deepfactors}
J.~Czarnowski, T.~Laidlow, R.~Clark, and A.~J. Davison, ``Deepfactors: Real-time probabilistic dense monocular slam,'' \emph{IEEE Robotics and Automation Letters}, vol.~5, no.~2, pp. 721--728, 2020.

\bibitem{teed2021droid}
Z.~Teed and J.~Deng, ``Droid-slam: Deep visual slam for monocular, stereo, and rgb-d cameras,'' \emph{Advances in neural information processing systems}, vol.~34, pp. 16\,558--16\,569, 2021.

\bibitem{eftekhar2021omnidata}
A.~Eftekhar, A.~Sax, J.~Malik, and A.~Zamir, ``Omnidata: A scalable pipeline for making multi-task mid-level vision datasets from 3d scans,'' in \emph{Proceedings of the IEEE/CVF International Conference on Computer Vision}, 2021, pp. 10\,786--10\,796.

\bibitem{burgdorfer2023v}
N.~Burgdorfer and P.~Mordohai, ``V-fuse: Volumetric depth map fusion with long-range constraints,'' in \emph{Proceedings of the IEEE/CVF International Conference on Computer Vision}, 2023, pp. 3449--3458.

\bibitem{song2021view}
S.~Song, D.~Kim, and S.~Choi, ``View path planning via online multiview stereo for 3-d modeling of large-scale structures,'' \emph{IEEE Transactions on Robotics}, vol.~38, no.~1, pp. 372--390, 2021.

\bibitem{hamdi2024ges}
A.~Hamdi, L.~Melas-Kyriazi, J.~Mai, G.~Qian, R.~Liu, C.~Vondrick, B.~Ghanem, and A.~Vedaldi, ``Ges: Generalized exponential splatting for efficient radiance field rendering,'' in \emph{Proceedings of the IEEE/CVF Conference on Computer Vision and Pattern Recognition}, 2024, pp. 19\,812--19\,822.

\bibitem{wang2004image}
Z.~Wang, A.~C. Bovik, H.~R. Sheikh, and E.~P. Simoncelli, ``Image quality assessment: from error visibility to structural similarity,'' \emph{IEEE transactions on image processing}, vol.~13, no.~4, pp. 600--612, 2004.

\bibitem{zhang2018unreasonable}
R.~Zhang, P.~Isola, A.~A. Efros, E.~Shechtman, and O.~Wang, ``The unreasonable effectiveness of deep features as a perceptual metric,'' in \emph{Proceedings of the IEEE conference on computer vision and pattern recognition}, 2018, pp. 586--595.

\end{thebibliography}

\end{document}